\documentclass[a4paper]{article}

\usepackage[english]{babel}
\usepackage[utf8]{inputenc}
\usepackage{amsmath}
\usepackage{graphicx}
\usepackage{hyperref}
\usepackage{wrapfig}
\usepackage{caption}
\usepackage{subcaption}

\title{Screen Tracking for Clinical Translation of Live Ultrasound Image Analysis Methods}

\author{Simona Treivase \and Alberto Gomez \and Jacqueline Matthew \and Emily Skelton \and Julia A. Schnabel \and Nicolas Toussaint\\
 King's College London, London, UK}
\date{}

\begin{document}

\maketitle

\begin{abstract}
Ultrasound (US) imaging is one of the most commonly used non-invasive imaging techniques. However, US image acquisition requires simultaneous guidance of the transducer and interpretation of images, which is a highly challenging task that requires years of training. Despite many recent developments in intra-examination US image analysis, the results are not easy to translate to a clinical setting. We propose a generic framework to extract the US images and superimpose the results of an analysis task, without any need for physical connection or alteration to the US system. The proposed method captures the US image by tracking the screen with a camera fixed at the sonographer's view point and reformats the captured image to the right aspect ratio, in 
87.66 $\pm$ 3.73\textit{ms} on average. 
It is hypothesized that this would enable to input 
such retrieved image into an image processing pipeline to extract information that can help improve the examination. 
This information could eventually be projected back to the sonographer's  field of view in real time using, for example, 
an augmented reality (AR) headset. 
\end{abstract}

\vspace{-3mm}
\section{Introduction}
\begin{figure}[tb]
\vspace{-5mm}
\centering
\includegraphics[scale=1.7]{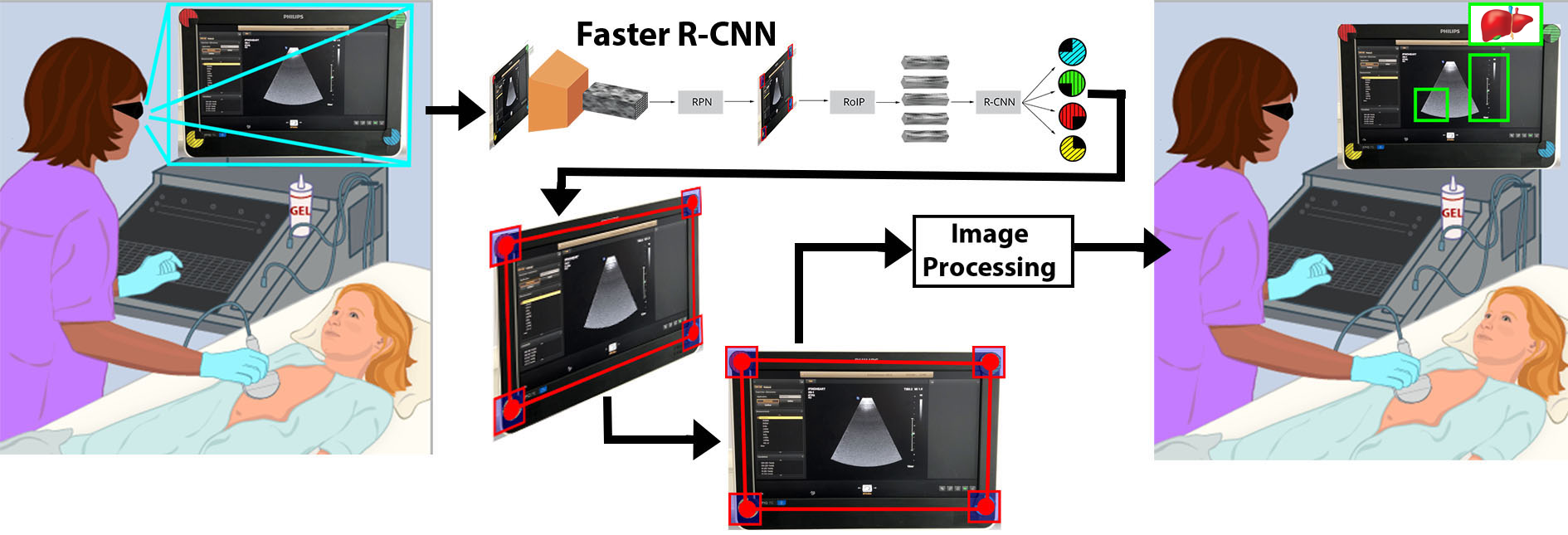}
\caption{Assisted scanning concept: screen capture  with a camera, geometric image reformatting, automated image processing and AR feedback during an Ultrasound examination.}
\label{fig:pipeline}
\vspace{-2mm}
\end{figure}

Recent developments in medical image processing and machine learning have begun to show great potential for real-time assisted US scanning, providing additional information about the scan, such as fetal organ classification and localization~\cite{toussaint2018weakly} or standard plane detection~\cite{7974824}. Typically, implementing such methods in a clinic would involve connecting the US system to a separate computer where the subsequent image processing takes place and where the any extracted measurements are shown, or modifying the US machine to process the images and display any measurements taken. The first alternative is impractical in the clinic, since the operator can not switch attention constantly between two screens (hand-eye synchronization is required to perform US). The second alternative requires access to the system and manufacturer-specific implementations which has specific commercial hurdles that limit the spread of the translation. Another (so far unexplored) alternative would be to capture the images from the US system, process them elsewhere, and overlay the results onto the screen of the US system, for example using augmented reality (AR) headsets. Towards this idea, in this paper, we propose to capture the images with a camera (hence removing the need for any manufacturer-specific adapter), and reformat the capture using the view parameters directly computed from the images to their original geometry for further processing. The contribution of this paper is two-fold: first, we introduce the concept of an end-to-end pipeline for assisted US examinations using AR (\ref{fig:pipeline}), and second we describe in detail the image capture component of this pipeline, which involves tracking the screen and unwrapping the images to their original geometry. 

\section{Methods}
Fig. \ref{fig:pipeline} illustrates the end-to-end assisted US concept. Image capture is summarised as follows: 1) a set of markers are attached to the US screen, 2) the screen is captured (with all markers in view), 3) Markers are detected and geometry is computed, and finally the image is warped back to the original position. The last step of the pipeline, which is projecting the results onto the screen using AR, is out of the scope of this paper and subject for future work. 

\textbf{(1) US screen tracking:} To retrieve the geometry of the screen within a video feed, we require the detection of four or more markers. We produced four physical markers with distinct colour and pattern to facilitate detection. We use Faster Region-based Convolutional Neural Networks (Faster R-CNN)~\cite{ren2015faster} for marker detection. This approach demonstrates good results and optimal trade-off between speed and accuracy~\cite{huang2017speed}. We chose MobileNet~\cite{DBLP:journals/corr/HowardZCKWWAA17} as feature extractor to prioritise inference speed, using weights pre-trained on ImageNet. We annotated a set of 250 images for training. The network converged in 20,000 iterations, using a fixed learning rate of 0.001 and a stochastic gradient descent optimiser. The network produces bounding boxes around detected markers.

\begin{figure}[b]
\centering
\vspace{-2mm}
\includegraphics[scale=0.43]{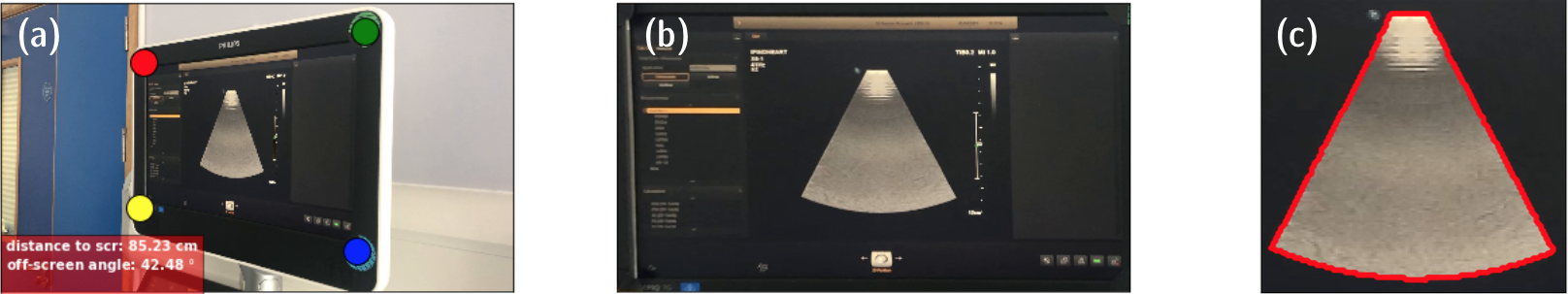}
\caption{(a) Detected US screen corners. (b) Inverse projective transformation $H^{-1}$ applied to (a). (c) US Frustum superimposed with edge detection result for Dice score analysis.}
\label{fig:projective-geometry-results}
\vspace{-2mm}
\end{figure}

\textbf{(2) Content extraction:} When all four markers are detected, central points of the bounding boxes are used to 
compute the projective transformation $H$~\cite{geetha2013automatic}. Its inverse $H^{-1}$ is applied to the distorted image and the original screen image is obtained. The results of such processing is shown in~Fig. \ref{fig:projective-geometry-results}.

\textbf{(3) Camera pose estimation:} The pose of the camera was obtained using the Direct Linear Transform solution followed by Levenberg-Marquardt optimization~\cite{levenberg1944method}. This is implemented in openCV in \textbf{solvePnP}. From the estimated pose we extract the distance to the screen and the off-screen angle.

\textbf{(4) Quantitative evaluation:} We used three metrics on a set of 296 annotated test images. First, the proportion of correctly detected markers $D$ is computed. Second, we extract the mean error in corners' localisation $l$ within each image. Third, when all four corners are detected, the US frustum from each distortion-corrected image (from \textbf{(2)}) is compared to that of the original image, using the Dice score as a metric.

\vspace{-3mm} 
\section{Results}

\begin{figure}[tb]
\centering
\vspace{-3mm}
\includegraphics[scale=0.43]{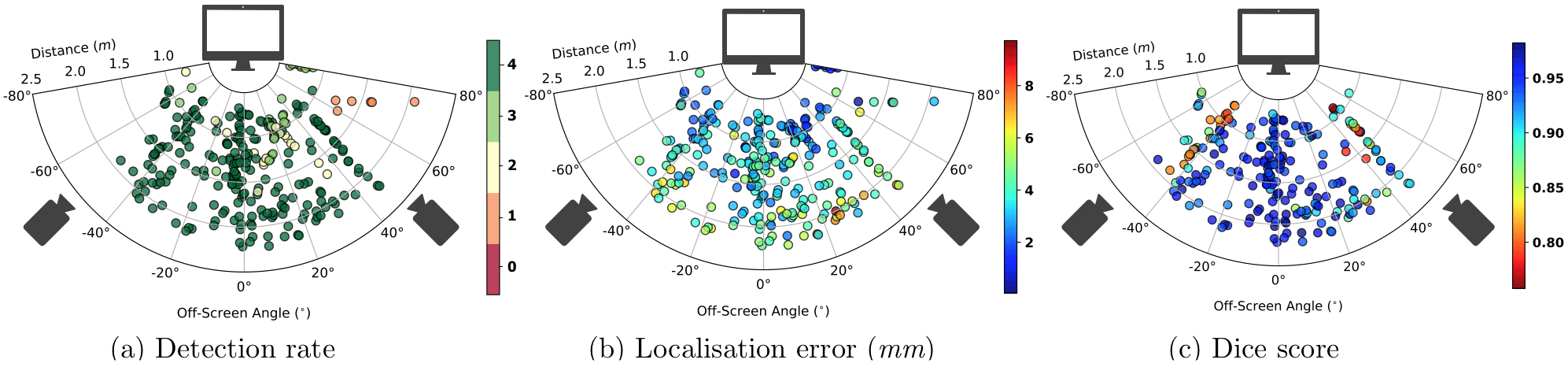}
\caption{Graphs representing the effects of distance and off-screen angle on (a) detection rate (number of correctly detected corners), (b) the corners' localisation error (in \textit{mm}), and (c) Dice score of the reconstructed US frustum w.r.t. ground truth.}
\label{fig:analysis-results}
\vspace{-2mm}
\end{figure}

Results of our quantitative analysis are summarised in~Fig. \ref{fig:analysis-results}. Detection rate is shown in Fig. \ref{fig:analysis-results}(a). 
Four markers were detected in 91\% of the test set, increasing to 98\% when the absolute off-screen angle did not exceed 60 degrees. The localisation error (Fig. \ref{fig:analysis-results}(b)) over the test set was 3.54 $\pm$ 1.52 \textit{mm} (mean $\pm$ standard deviation) and does not demonstrate strong correlation with the off-screen angle, however predictably worsens as the distance increases. The results in Fig. ~\ref{fig:analysis-results}(c) show the frustum Dice score metric with an average of 0.92 $\pm$ 0.05. One side notably performs worse than the other side, due to strong reflections from the window. From these preliminary results, our approach appears robust to a range of off-screen angles and distances that would normally occur in the clinic.

\vspace{-3mm}
\section{Conclusion and Future Work}

We presented a pipeline for screen tracking and content extraction for US clinical translation. The proposed methods approaches real time at 11 images per second (laptop with NVIDIA GTX960). This will be further extended in future work to extract relevant information from the retrieved US image and project it back to the world using an AR headset. The opportunities are not limited to US analysis and can be extended to other imaging modalities that have real-time imaging capabilities. The potential of such methods may improve scanning times and quality of US screening programs. 

\section*{Acknowledgements}
This work was supported by the Wellcome Trust IEH Award [102431]. This work was supported by the Wellcome/EPSRC Centre for Medical Engineering at King’s College London [WT 203148/Z/16/Z]. The research was funded/supported by the National Institute for Health Research (NIHR) comprehensive Biomedical Research Centre awarded to Guy's and St Thomas' NHS Foundation Trust and King's College London. The views expressed are those of the author(s) and not necessarily those of the NHS, the NIHR or the Department of Health.

\bibliography{bibliography.bib}
\bibliographystyle{plain}

\end{document}